\pdfoutput=1

\documentclass[11pt]{article}

\usepackage[]{coling}

\usepackage{times}
\usepackage{amssymb}
\usepackage{latexsym}
\usepackage{paralist}
\usepackage{graphicx}
\usepackage{hyperref}
\usepackage{amsmath}
\usepackage{url}

\usepackage{bbding}
\usepackage[T1]{fontenc}

\usepackage[utf8]{inputenc}

\usepackage{microtype}

\usepackage{inconsolata}
\usepackage{color}

\title{QCRD: Quality-guided Contrastive Rationale Distillation for Large Language Models}
\author{
    \textbf{Wei Wang\textsuperscript{\rm 1,2}},
    ~\textbf{Zhaowei Li\textsuperscript{\rm 2,3}},
    ~\textbf{Qi Xu\textsuperscript{\rm 2}},
    ~\textbf{Yiqing Cai\textsuperscript{\rm 2}},
    ~\textbf{Hang Song\textsuperscript{\rm 2}},\\
    ~\textbf{Qi Qi\textsuperscript{\rm 2}},
    ~\textbf{Ran Zhou\textsuperscript{\rm 2}},
    ~\textbf{Zhida Huang\textsuperscript{\rm 2}},
    ~\textbf{Tao Wang\textsuperscript{\rm 2}},
    ~\textbf{Li Xiao\textsuperscript{\rm 1}}\\
    \textsuperscript{\rm 1}University of Science and Technology of China, \textsuperscript{\rm 2}ByteDance Inc, \textsuperscript{\rm 3}Fudan University \\
    {\tt wangweiii@mail.ustc.edu.cn} \\
}

\begin{document}

\maketitle
\begin{abstract}
The deployment of large language models (LLMs) faces considerable challenges concerning resource constraints and inference efficiency. Recent research has increasingly focused on smaller, task-specific models enhanced by distilling knowledge from LLMs. However, prior studies have often overlooked the diversity and quality of knowledge, especially the untapped potential of negative knowledge. Constructing effective negative knowledge remains severely understudied. In this paper, we introduce a novel framework called quality-guided contrastive rationale distillation aimed at enhancing reasoning capabilities through contrastive knowledge learning. For positive knowledge, we enrich its diversity through temperature sampling and employ self-consistency for further denoising and refinement. For negative knowledge, we propose an innovative self-adversarial approach that generates low-quality rationales by sampling previous iterations of smaller language models, embracing the idea that one can learn from one’s own weaknesses. A contrastive loss is developed to distill both positive and negative knowledge into smaller language models, where an online-updating discriminator is integrated to assess qualities of rationales and assign them appropriate weights, optimizing the training process. Through extensive experiments across multiple reasoning tasks, we demonstrate that our method consistently outperforms existing distillation techniques, yielding higher-quality rationales. Our codes will be released soon.
\end{abstract}

\section{Introduction}
The reasoning capabilities of large language models (LLMs) have been observed to scale their model sizes, while necessitating substantial memory and computing resources~\cite{chowdhery2023palm, wei2022emergent}. As such, efficient model compression is crucial in the deployment of LLMs, especially on resource-limited devices or platforms. Knowledge distillation from an LLM (teacher) to a smaller, more manageable language model (student) has recently emerged as a powerful and promising technique for model compression~\cite{hinton2015distilling, phuong2019towards}. However, it is still open how to best reduce the performance gap between the teacher and the student on complex reasoning tasks~\cite{zelikman2022star}.

\begin{figure}[t]
\centering
\includegraphics[width=1\columnwidth]{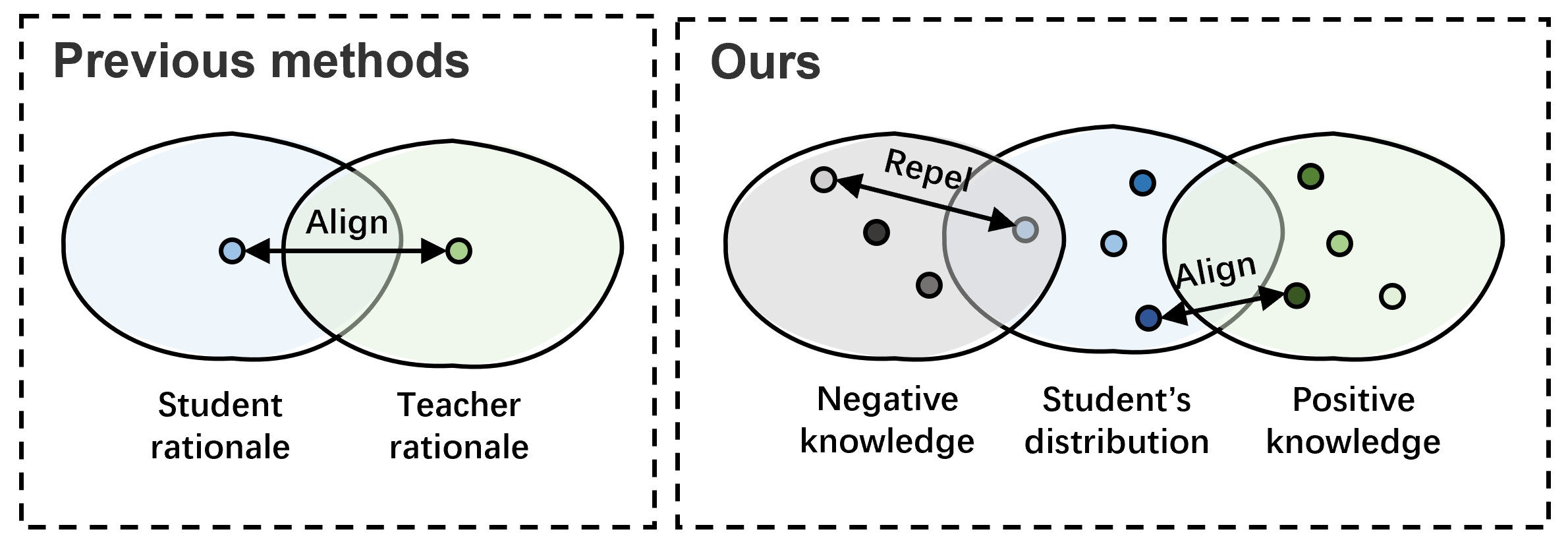}\\
\caption{Comparison between previous methods and our proposed method, where circle points denote rationales, and colors of the circle points correspond to rationale types, and shades of darker indicate higher qualities. The "align" means minimizing the distance between rationales, while the "repel" means maximizing the distance.}\label{comparison}
\end{figure}

In this regard, it has more recently been shown that adding explanation-augmented prompts, especially, Chain-of-Thought (CoT)~\cite{wei2022chain},
can enable \textcolor{black}{LLMs} to generate reasonable explanations (also referred to as rationales) to justify the reasoning outcomes~\cite{li2022explanations}. Distilling these rationales into smaller language models has been demonstrated to effectively improve the overall performance~\cite{hsieh2023distilling, li2022explanations}.
For example, distilling Step-by-Step (DSS)~\cite{hsieh2023distilling} was proposed as an innovative CoT distillation approach, which employed rationales from an LLM to guide a smaller language model under a multi-task learning setting. It involved training the smaller language model simultaneously on both label prediction and rationale generation tasks, effectively leveraging their mutual benefits.

The essence of such distilling rationales is to guide the model in learning additional knowledge related to the labels. Knowledge can be generally concluded into two classes: positive and negative. Previous works on rationale distillation, although effective, still suffer from certain drawbacks. On the one hand, positive knowledge for distillation may be limited and noisy. Methods~\cite{fu2023specializing, hsieh2023distilling, magister2022teaching,chen2024learning} treated rationales generated by LLMs as golden answers and aimed to minimize the gap between these rationales and those generated by smaller language models. However, despite LLMs' powerful zero-shot/few-shot abilities, they may occasionally produce incorrect reasoning steps, leading to erroneous rationales/answers. Such erroneous rationales may degrade the reasoning performance of the distilled smaller language models. On the other hand, generating negative rationales and incorporating them into CoT distillation remain understudied, while negative knowledge has early proved constructive and effective for models.

To this end, we propose a general method, named Quality-guided Contrastive Rationale Distillation (QCRD), to guide the knowledge distillation to smaller language models from a contrastive learning perspective. The comparison between the previous methods and our proposed QCRD is illustrated in Fig.~\ref{comparison}. Specifically, the previous methods focus on the alignment between the rationale of the student model and the corresponding one of the teacher model, while our proposed QCRD aligns the student's distribution and contrastive knowledge distribution with various sampled rationales. The core design of QCRD is to generate a diverse set of contrastive rationales and efficiently distill them into student models. For the positive part, to ensure the quality and variety of positive rationales, we prompt the LLM and sample the output to generate multi-round rationales for each input question. We then apply the self-consistency to denoise the rationale set and split it into positive rationales and negative rationales. For the negative part, we employ a self-adversarial strategy inspired by~\cite{silver2018general} during training to generate low-quality rationales from previous iterations of smaller language models with a high sampling temperature and treat them as negative rationales. Finally, for better knowledge learning, we present a contrastive loss to distill both positive and negative rationales into smaller language models. A discriminator is adopted to assess the qualities of the rationales and assign them appropriate weights to optimize the training process across the datasets. 

To demonstrate the superiority of QCRD, we conduct comprehensive experiments with two smaller types of T5 models~\cite{raffel2020exploring}, i.e., T5-base (220M parameters) and T5-small (60M parameters), on four popular datasets, followed by detailed analysis and discussion. Our main contributions of this paper can be summarized below.
\begin{itemize}
\item We first develop a general CoT distillation approach (i.e., QCRD) from a contrastive learning perspective, aiming to guide the student model to learn both positive and negative knowledge from rationales.
\item We explore a contrastive distillation loss to facilitate effective distillation of the generated positive and negative rationales, where the qualities of the rationales judged by a discriminator are considered to optimize the training process across the whole datasets.
\item Experimental results across multiple datasets show that QCRD demonstrably outperforms existing benchmarks, evidencing its efficiency in transferring contrastive reasoning knowledge to the smaller language model.
\end{itemize}

\begin{figure*}[!t]
\centering
\includegraphics[width=2\columnwidth]{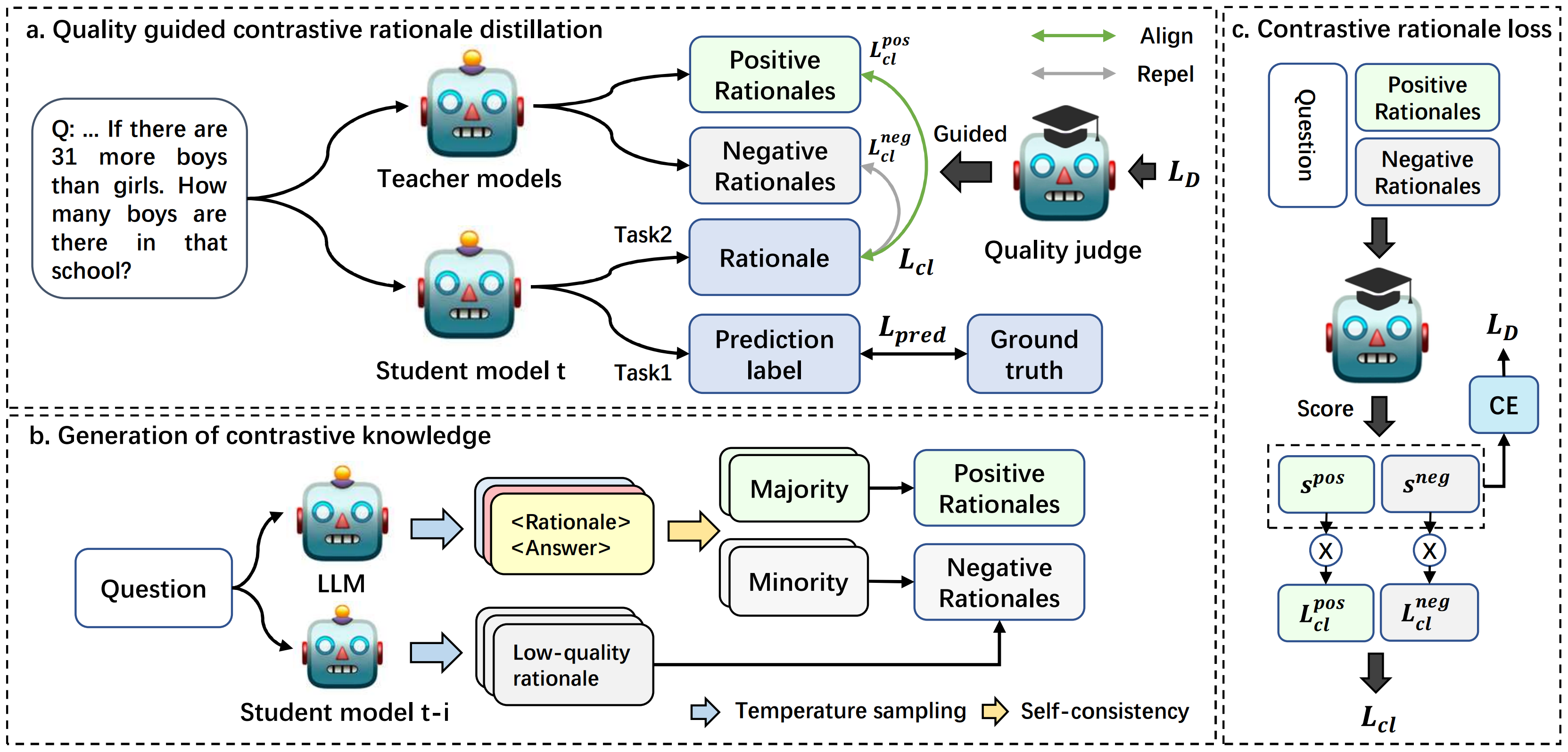}\\
\caption{Illustration of the proposed quality-guided contrastive rationale distillation for distilling contrastive knowledge from teacher models into the student model. Fig.a represents our multi-task framework, i.e., the main prediction label task and additional rationale task. Fig.b represents generation of contrastive rationales for distillation. Fig.c represents details about the quality-guided contrastive rationale loss, and CE denotes the cross-entropy.}
\label{QCRD}
\end{figure*}

\section{Related Work}
\noindent\textbf{Knowledge distillation from LLMs.} Knowledge distillation (KD) is a highly effective technique for transferring knowledge from larger teacher models to smaller student models that are more suitable for practical applications~\cite{fu2023specializing, hsieh2023distilling, magister2022teaching, chen2024learning, wang2023scott}. The KD technique can be generally classified into two different categories: (1) Black-box KD: only the teacher’s predictions are accessible; (2) White-box KD: it provides access to the teacher's parameters. Both of them have shown promising potential in fine-tuning smaller models on the prompt response pairs generated by LLMs~\cite{zhu2023survey}. In this paper, we hypothesize that only the predictions (predict labels and rationales) generated by LLMs are accessible.

\noindent\textbf{Multi-task learning with LLM generated rationales}. Current LLMs have already exhibited their capabilities to generate high-quality reasoning steps, resulting in rationales of their predictions~\cite{kojima2022large}, and these rationales have been found to be valuable additional knowledge for fine-tuning smaller models~\cite{hsieh2023distilling}. A multi-task learning framework is commonly employed that enforces smaller models to output corresponding rationales, while maintaining their original functionality. However, previous studies only focused on aligning the output of the smaller model with that of the LLM with a single loss form~\cite{hsieh2023distilling, magister2022teaching}.

\noindent\textbf{Self-consistency of LLMs}. The self-consistency of LLMs refers to the capacity to maintain coherent and rational reasoning during input processing. Based on the intuition that complex reasoning tasks typically admit multiple reasoning paths that reach a correct answer, the self-consistency can improve the LLMs' reasoning performance by integratedly sampling CoT outputs several times and choosing the most consistent predict answer~\cite{stanovich199124, wang2022self}.

\noindent\textbf{Contrastive learning for LLMs}. Contrastive learning has demonstrated its efficiency across diverse domains, e.g., computer vision, natural language processing~\cite{jaiswal2020survey, le2020contrastive}. Notably, the application of contrastive learning to LLMs has recently emerged, highlighting the effectiveness of incorporating negative knowledge implicitly in model's inputs and showing promising outcomes~\cite{li2024turning,chen2024magdi}. However, to the best of our knowledge, the application of contrastive learning in CoT rationale distillation has not been explored thus far.

\section{Methodology}
\label{method}
In this paper, we first propose a general contrastive CoT distillation approach, called quality-guided contrastive rationale distillation (QCRD), for training smaller models by distilling contrastive knowledge from teacher models. As illustrated in Fig.~\ref{QCRD}, our approach consists of the following three parts. (1) Following the method developed in \cite{hsieh2023distilling}, we apply a multi-task learning framework for the supervised training of the student model, i.e., the main prediction label task and additional rationale generation task; see Fig.~\ref{QCRD}a. (2) As displayed in Fig.~\ref{QCRD}b, we design a general approach to generate contrastive knowledge from LLMs and student model itself for rationale distillation. (3) As shown in Fig.~\ref{QCRD}c, for better knowledge learning from rationales, we design a quality-guided contrastive learning strategy, where a contrastive loss is applied with the guidance of an online-updated discriminator to distinguish between positive and negative rationales and assign them quality scores.

\subsection{Multi-task learning framework for the student model}
Previous works have already demonstrated the advantages of the multi-task learning framework~\cite{fu2023specializing, hsieh2023distilling, magister2022teaching}. Accordingly, as shown in Fig.~\ref{QCRD}a, we apply the label prediction task and the rationale generation task to the training of smaller language models. Specifically, we use different prefixes to enforce smaller language models to generate different types of output. Given an input question, for the label prediction task, the smaller language model outputs the prediction label with input prefix $<Predict>$, while for the rationale generation task, it outputs the corresponding explanation with input prefix $<Explain>$. These outputs are then aligned to the corresponding ground truth and rationales using autoregressive loss, respectively.

\subsection{Generation of contrastive knowledge}
\begin{figure}[!ht]\small
\centering
\includegraphics[width=1\columnwidth]{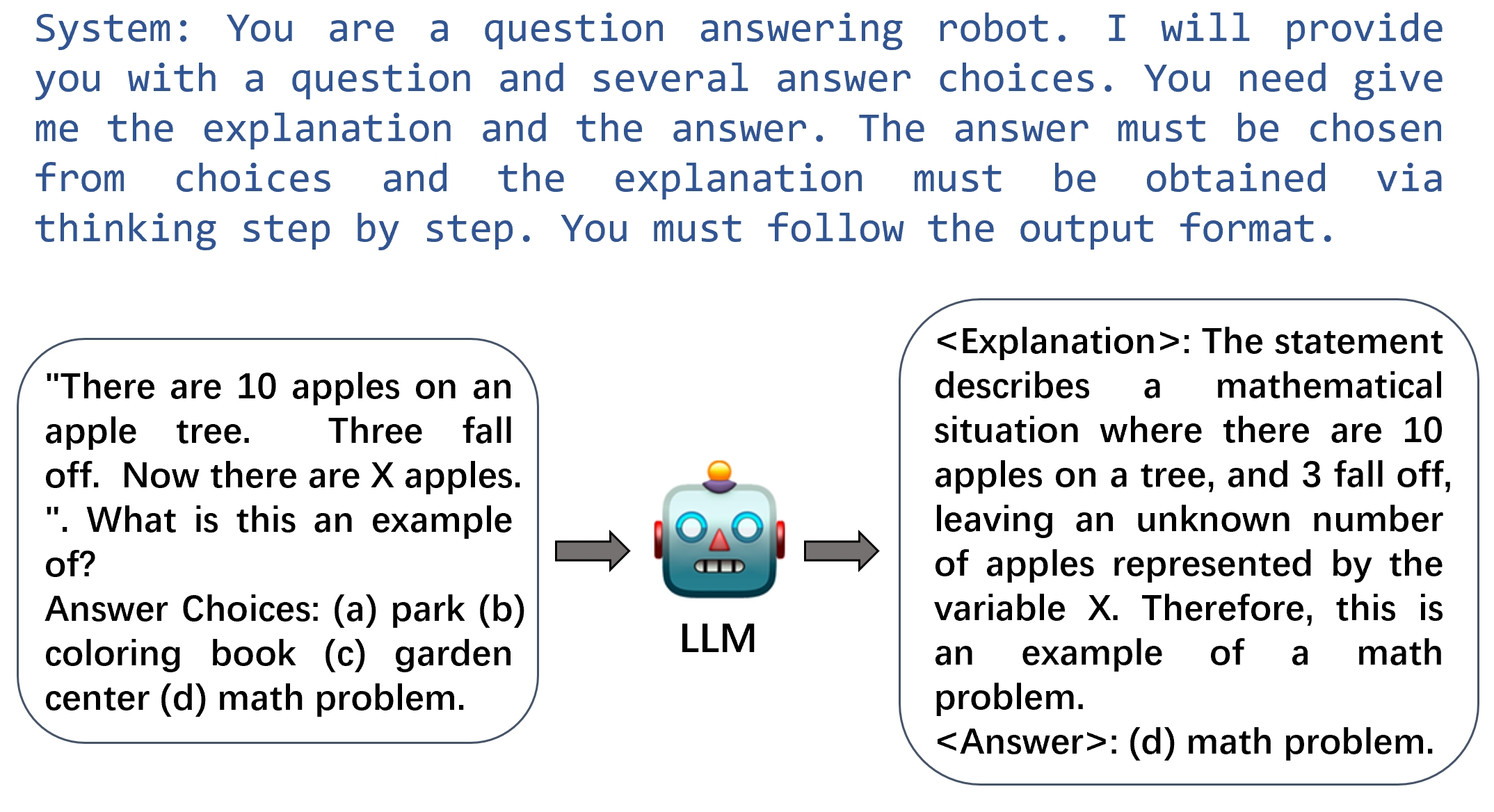}\\
\caption{A case of the prompt and rationale output.}\label{example}
\end{figure}

We use CoT prompting~\cite{wei2022chain} to elicit and extract rationales from LLMs. As illustrated in Fig.~\ref{example}, the LLM is provided with few examples to follow the output format. Instead of only generating one output for each input, we replace the “greedy decode” in CoT prompting with sampling from the language model's decoder to generate a diverse set of reasoning paths~\cite{wang2022self}. We apply temperature sampling~\cite{renze2024effect} to the LLM $K$ times, where a temperature value $\tau$ can control the diversity of the generated output. Therefore, for each input, there are $K$ pairs of rationales and corresponding labels.

\subsubsection{Build positive and negative rationale sets}\label{set}
Language models are not perfect reasoners in the sense that they might occasionally produce an incorrect reasoning path or make a mistake in one of the reasoning steps. In general, evidence shows that correct reasoning processes, even though they are diverse, tend to have greater agreement in their final answers than incorrect reasoning processes~\cite{wang2022self}. We thus choose the rationales with the most consistent label in all outputs of the LLM as positive rationales, and the remaining rationales are viewed as negative rationales. However, the sampling times mainly influence the proportion of positive and negative samples (the number of negative samples is always smaller, and detailed statistical descriptions can be found in Appendix \ref{annotations}), along with the time and storage costs. Moreover, these negative rationales from LLMs are likely to be positive for smaller language models, which may limit their effectiveness.

To deal with this issue, we conduct a self-adversarial mechanism that the student model generates its own negative rationales by sampling from its previous iterations with a high temperature value during training (we illustrate its rationality in Section~\ref{temperature_sampling} and demonstrate its superiority in Appendix~\ref{Iteration-before-models}), and we regard these low-quality rationales as negative ones based on the hypothesis that the rationale quality of LLMs is higher than that of smaller models. As a result, for each input question $\mathbf{x}=[x_{1}, x_{2}, ..., x_{n}]$, we collect a positive rationale set $S_{pos}=\{\mathbf{r}_{1}^{pos}, \mathbf{r}_{2}^{pos}, ..., \mathbf{r}_{m}^{pos}\}$ and a negative rationale set $S_{neg}=\{\mathbf{r}_{1}^{neg}, \mathbf{r}_{2}^{neg}, ..., \mathbf{r}_{k}^{neg}\}$.

\subsection{Contrastive knowledge distillation}
In this subsection, we present our designed quality-guided contrastive rationale distillation for better knowledge learning.
\subsubsection{Train a discriminator to judge rationales}\label{Discriminator}
The quality of rationales for the same question still differs. Moreover, as the training epoch increases, the rationales generated by the above self-play may become gradually closer to the positive rationales, and then viewing them as negative ones is no longer reasonable. Therefore, there is a need to train a discriminator $\mathcal{D}$ that can effectively judge the positive and negative rationales and output a score that represents the quality of each rationale. The input of the discriminator $\mathcal{D}$ is the question and the rationale, and we take an encoder architecture to measure the score, i.e.,
\begin{equation}\label{discriminator}\small
s_{j}^{pos} = \mathcal{D}\left(\mathbf{x}, \mathbf{r}_{j}^{pos}\right) \;\;\mbox{or}\;\; s_{j}^{neg} = \mathcal{D}\left(\mathbf{x}, \mathbf{r}_{j}^{neg}\right).
\end{equation}
We pretrain the $\mathcal{D}$ with the positive and negative rationales from the LLM, and during training, the discriminator $\mathcal{D}$ is updated at regular epoch intervals (details can be seen in Appendix \ref{Judge}). The loss function can be formulated as
\begin{equation}\label{discriminatorloss}\small
\mathcal{L}_{\mathcal{D}}=\mathbb{E}_{\mathbf{x}}\left[-\log\frac{\sum_{j=1}^{m}\exp(s^{pos}_{j})}{\sum_{j=1}^{k}\exp(s^{neg}_{j})}\right].
\end{equation}

\subsubsection{Quality-guided contrastive distillation}
\label{kd}
As mentioned in sec.~\ref{set}, there is a diverse set of positive rationales. In addition, the negative rationales are of significance, which can enforce the smaller model away from their distribution. Since some of the negative samples are generated by the previous-iteration smaller model, the smaller model can further refine its reasoning capability through playing against instances of itself and promote the generated rationales closer to golden rationales of the LLM. Therefore, we propose a many-to-one contrastive distillation loss, while previous studies typically utilize a single rationale for each question and distill it into the smaller model, i.e.,
\begin{equation}\label{loss}\small
\mathcal{L}_{cl} = \frac{1}{N}\sum_{i=1}^{N}[l(f(\mathbf{x}_{i}),S_{pos}^i)-\beta \cdot l(f(\mathbf{x}_{i}),S_{neg}^i)],
\end{equation}
where $\mathbf{x}_{i}$ denotes the $i$-th question, $S_{pos}^{i}$ and $S_{neg}^{i}$ denote the corresponding positive and negative rationale sets, respectively, $N$ is the number of questions, and $\beta>0$ is a tunable hyper-parameter. The function $l(\cdot)$ denotes the cross-entropy loss and $f(\cdot)$ denotes the rationale generation for its given input. In (\ref{loss}),
\begin{equation}\label{min}\small
l(f(\mathbf{x}_{i}),S_{pos}^{i})=\mathop{\min}_{\mathbf{r}_{j}^{pos,i}\in S_{pos}^{i}}\left\{ l\left(f(\mathbf{x}_{i}),\mathbf{r}_{j}^{pos,i}\right)\right\},
\end{equation}
\begin{equation}\label{max}\small
l(f(\mathbf{x}_{i}),S_{neg}^{i})=\mathop{\max}_{\mathbf{r}_{j}^{neg,i}\in S_{neg}^{i}} \left\{l\left(f(\mathbf{x}_{i}),\mathbf{r}_{j}^{neg,i}\right)\right\},
\end{equation}
which are motivated by the sequence-level distillation~\cite{kim2016sequence}. 
 Moreover, we set a margin $\delta$ for the negative rationales to filter out cases that are too simplistic, i.e., $l(f(\mathbf{x}),\mathbf{r}_{j}^{neg})=\min(l(f(\mathbf{x}),\mathbf{r}_{j}^{neg}) - \delta, 0)$ with respect to the $j$-th negative rationale for an input question $\mathbf{x}$. 
 Let us rethink the effectiveness of negative rationales generated by the previous-iteration smaller model, which enforces the smaller model to break out of local optima and yield a golden rationale that is closer to the output of the LLM. However, when the smaller model comes to converging, the previous-iteration smaller model is likely to output the rationales that are similar to those of the LLM, and then regarding them as negative samples is inaccurate. To address this issue, we introduce the quality-guided distillation to optimize the training process and redefine the loss formulas in (\ref{min}) and (\ref{max}) as, respectively,
\begin{equation}\label{confidence1}\small
l(f(\mathbf{x}_{i}),S_{pos}^{i})
= s^{pos,i} \cdot \mathop{\min}_{\mathbf{r}_{j}^{pos,i}\in S_{pos}^{i}}\left\{ l\left(f(\mathbf{x}_{i}),\mathbf{r}_{j}^{pos,i}\right)\right\},
\end{equation}
\begin{equation}\label{confidence1}\small
l(f(\mathbf{x}_{i}),S_{neg}^{i})
=(1-s^{neg,i}) \cdot \mathop{\max}_{\mathbf{r}_{j}^{neg,i}\in S_{neg}^{i}} \left\{l\left(f(\mathbf{x}_{i}),\mathbf{r}_{j}^{neg,i}\right)\right\},
\end{equation}
where $s^{pos,i}$ and $s^{neg,i}$ are the corresponding quality scores obtained by the discriminator $\mathcal{D}$. By (6) and (7), the positive rationales of higher quality should have larger weights across the datasets, while for the negative rationales of higher quality, it is on the contradiction. In the latter sec.~\ref{many-to-one}, we will further discuss different schemes for the many-to-one distillation.

\subsubsection{Training loss}
The \textcolor{black}{final} training loss is given by
\begin{equation}\label{lossfunctiom}
\mathcal{L}_{total}=\alpha_{1}\mathcal{L}_{pred}+\alpha_{2}\mathcal{L}_{cl}+\alpha_{3}\mathcal{L}_{\mathcal{D}},
\end{equation}
where \textcolor{black}{$\left\{\alpha_{i}\right\}_{i=1}^3>0$ are tunable hyper-parameters}, $\mathcal{L}_{pred}$ represents the cross entropy loss of the label prediction task, $\mathcal{L}_{cl}$ is \textcolor{black}{the many-to-one contrastive distillation loss in (\ref{loss}), and $\mathcal{L}_{\mathcal{D}}$ is the discriminator loss in (\ref{discriminatorloss})}.

\section{Experiments}
\subsection{Experimental setting}
\textbf{Datasets}. We conducted extensive experiments on four widely-used benchmark datasets (see details in Appendix Table~\ref{datasets}) across three different natural language processing tasks, including SVAMP~\cite{patel2021nlp} for arithmetic word problem solving, CQA~\cite{talmor2018commonsenseqa} for commonsense question answering, as well as e-SNLI~\cite{camburu2018snli} and ANLI~\cite{nie2019adversarial} for natural language inference. The rationales we used were generated by GPT-3.5-turbo\footnote{\url{https://platform.openai.com/docs/models}} and an opened code source by~\cite{hsieh2023distilling} was referred.

\noindent\textbf{Implementation details}. Following the properties of CoT and comparative experimental studies in~\cite{hsieh2023distilling}, our \textcolor{black}{QCRD} adopted T5-base (\textcolor{black}{220M parameters}) and T5-small (\textcolor{black}{60M parameters}) \textcolor{black}{as} the student model, respectively. $\alpha_{1}$, $\alpha_{2}$, $\alpha_{3}$ were set to $0.5$ empirically. $\alpha_{3}$ was multiplied by $0.9$ per iteration. We set $\beta=0.2$ and $\delta=3$. We sampled the LLM's output $5$ times with the temperature being $0.7$, and sampled $5$-iteration-before models with the temperature being $1.5$. The batchsize was 8 and learning rate was 1e-5. We trained our models with 10000 max steps on one A100-80G about 13 hours for T5-base and 8.5 hours for T5-small. The reported metric was accuracy.

\noindent\textbf{Baselines}. Four methods in learning task-specific models \textcolor{black}{were} compared, \textcolor{black}{i.e.,} (1) Finetuning, which is the standard finetuning with the prevailing pretrain-then-finetune paradigm that finetunes a model with ground-truth labels via standard label supervision~\cite{howard2018universal}; (2) Single-Task, where student models are distilled to predict labels with the teacher model’s predicted labels; (3) DSS~\cite{hsieh2023distilling}, where student models are distilled with both the predict labels and rationales of the LLM; (4) Mutual information (MI)~\cite{chen2024learning}, which is based on DSS and applies an additional task to maximizing the mutual information between prediction labels and rationales.

\begin{figure*}[!ht]
\centering
\includegraphics[width=2\columnwidth]{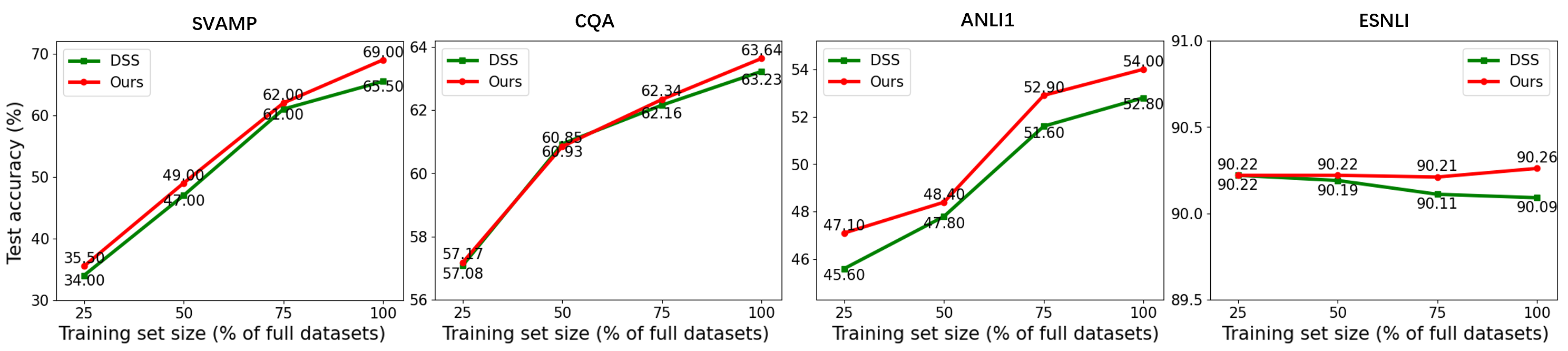}\\
\caption{Comparisons with varying sizes of training datasets on the T5-base model for four benchmarks. }\label{size}
\end{figure*}

\subsection{\textcolor{black}{Experimental results}}

\noindent\textbf{Experiments across four benchmarks}. We conducted experiments across four benchmarks with two types of T5-model to evaluate the effectiveness of our proposed method. In the top of Table~\ref{table1}, we summarized the experimental results of the T5-base model distilled by our method and the baselines individually on all the four datasets. \textcolor{black}{Of note, in Single-Task}, the rationale and label \textcolor{black}{were} combined into a single sequence, which was then treated as the target \textcolor{black}{during the training}~\cite{hsieh2023distilling}. It is clear that our method outperformed the baselines on most datasets, particularly when compared to the baseline DSS.

\begin{table}[!h]\small
    \centering
        \renewcommand\arraystretch{1.15}
    \caption{CoT distillation results on the T5-base model.}
    \begin{tabular}{ccccc}
    \hline
        Model & SVAMP & CQA & ANLI1 & ESNLI \\ \hline
        \textcolor{black}{Finetuning} & 63.0 & 62.19  & 43.58 & 88.38 \\
        Single-Task & 59.0 & 63.11 & 47.90 & 88.77 \\
        DSS & 65.5 & 63.23 & 52.80 & 90.09 \\
        MI & 67.5 & 63.50 & \textbf{54.20} & 90.15 \\
        Ours & \textbf{69.0} & \textbf{63.64} & 54.00 & \textbf{90.26} \\ \hline

    \end{tabular}
    \label{table1}
\end{table}

In like manner, we performed our method and the baselines individually on the T5-small model, and their performance on all the four datasets was presented in Table~\ref{table2}. Our method consistently surpassed the baselines on all the four datasets.

\begin{table}[!h]\small
    \renewcommand\arraystretch{1.1}
    \centering
    \caption{CoT distillation results on \textcolor{black}{the} T5-small model.}
    \begin{tabular}{ccccc}
    \hline
        Model & SVAMP & CQA & ANLI1 & ESNLI \\ \hline
        Finetuning & 45.00 & 43.16 & 42.00 & 82.90  \\
        Single-Task & 46.50 & 44.98 & 42.50 & 83.67 \\
        DSS & 48.00 & 45.21 & 42.80 & 84.23 \\
        MI & 47.00 & 45.49 & 42.10 & 83.55 \\
        Ours & \textbf{50.50} & \textbf{46.11} & \textbf{44.10} & \textbf{85.30} \\ \hline
    \end{tabular}
    \label{table2}
\end{table}

\noindent\textbf{Distillation with LLM labels}. To evaluate the impact of label qualities on CoT distillation, without loss of generality, we conducted additional experiments on the three datasets (namely, CQA, ANLI1, and ESNLI) using the T5-base model distilled by our method and DSS. Instead of using ground truth labels, we employed \textcolor{black}{the} labels generated by GPT-3.5-turbo to distill student models. The results were presented in Table~\ref{table3}. On one hand, from the top of Table~\ref{table3}, it demonstrates the effectiveness of temperature sampling and self-consistency (SC), which help denoise rationales and their corresponding labels. On the other hand, the results at the bottom of Table~\ref{table3} indicate that our method outperformed DSS on CQA and ANLI1, even when utilizing labels generated by the LLM.

\begin{table}[!ht]\small
    \renewcommand\arraystretch{1.15}
    \centering
    \caption{CoT distillation results on the T5-base model using predicted labels (noisy labels) from the LLM.}
    \begin{tabular}{ccccc}
    \hline
        Model & CQA & ANLI1 & ESNLI \\ \hline
        GPT 3.5 & 66.30 & 78.21 & 66.27 \\
        GPT 3.5 with SC & 69.05 & 80.15 & 67.08 \\ \hline
        DSS & 59.15 & 44.10 & \textbf{74.88} \\
        MI & 59.22 & 45.90 & 74.67 \\
        Ours & \textbf{59.80} & \textbf{46.70} & \textbf{74.88} \\ \hline
    \end{tabular}
    \label{table3}
\end{table}

\noindent\textbf{Distillation with smaller datasets}. \textcolor{black}{In addition, to demonstrate the superiority of our method on smaller datasets, we compared the performance of DSS and our method using T5-base models on varying sizes of each of the four datasets. Fig.~\ref{size} illustrates that our method achieved better performance in most cases}. It indicates the robustness and generality of our proposed QCRD.

\noindent\textbf{Ablation study on QCRD}. Compared to previous related methods, the contrastive distillation in our QCRD introduces several key enhancements as follow. (1) The extension and denoising for positive knowledge (ED): we sample the outputs of the LLM and leverage the self-consistency to denoise rationales. (2) The distillation for negative knowledge (NK): we incorporate a self-supervised mechanism to generate low-quality rationales as negative rationales. (3) The guidance of the Quality Judge (QJ): the use of discriminator helps assess rationales and optimize the training process. Additional experiments were so conducted on SVAMP to evaluate the effectiveness of each module, with the results being summarized in Table~\ref{table4}. The findings demonstrated that integrating more high-quality rationales significantly improved performance, while the inclusion of negative rationales proved effective. The discriminator mechanism played a positive role by considering the quality of each rationale, and we further found that the results when using the Quality Judge were more stable.

\begin{table}[!ht]\small
    \renewcommand\arraystretch{1.1}
    \centering
    \caption{Ablation study on T5-base model, where ED denotes positive knowledge extension and denoising, NK denotes negative knowledge, and QJ denotes using of Quality Judge.}
    \begin{tabular}{cccc}
    \hline
        w/o ED & w/o NK & w/o QJ & SVAMP
 \\
        \XSolidBrush & \XSolidBrush & \XSolidBrush & 65.5 \\
        \Checkmark & \XSolidBrush & \XSolidBrush & 67.0 \\
        \Checkmark & \Checkmark & \XSolidBrush & 68.5 \\
        \Checkmark & \Checkmark & \Checkmark & 69.0 \\ \hline
    \end{tabular}
    \label{table4}
\end{table}

\section{Discussion}\label{discussion}
\subsection{Different contrastive distillation schemes} \label{many-to-one}
In sec.~\ref{kd}, We defined the many-to-one distillation by taking the min loss for positive rationales and max loss for negative rationales (i.e., MinMax), which imposes a relatively weak constraint on rationale alignment. We further discuss different schemes for the many-to-one distillation. (1) MaxMin: we compute the max loss for positive rationales and min loss for negative rationales. This scheme enforces the smaller model to learn hard rationale examples. (2) Sampling: we randomly choose a positive rationale and a negative rationale for each input. (3) Mean: we average the loss for all rationales. (4) Weighted mean (W-mean): we weight the loss with quality scores and then average the loss. The results of the T5-base model distilled by our method on SVAMP were presented in Table~\ref{methods} with respect to the above different schemes. One can clearly see that the MinMax achieved the best performance. Besides, the Mean scheme had a negative impact on the results. The reason may be that enforcing small models align with multi-target rationales of differences is not suitable, especially for positive knowledge.

\begin{table}[!ht]\small
    \renewcommand\arraystretch{1.4}
    \setlength{\tabcolsep}{1.2mm}{
    \centering
    \caption{Results of our method with different many-to-one distillation schemes on SVAMP.}
    \begin{tabular}{cccccc}
    \hline
        Model & MinMax & MaxMin & Sampling & Mean & W-mean \\ \hline
        T5-base & 69.0 & 67.0 & 66.5 & 65.0 & 66.0 \\ \hline
    \end{tabular}
    \label{methods}}
\end{table}

\subsection{Influence of the sampling count}
In the above experiments, we sampled the output of the LLM five times and the output of iteration-before model once. We further explore the influence of the sampling count. When fixing the sampling counts for iteration-before models, results of setting different sampling counts for the LLM on SVAMP were displayed in the top of Table~\ref{table6}. Moreover, when fixing the sampling counts for the LLM, results on SVAMP were displayed in the bottom of Table~\ref{table6} in terms of different numbers of generated negative samples. We found that sampling many negative rationales had an adverse impact on the performance, and the best performance was achieved when $k$ was 1. Note that when $m=1$, the performance of our method was still better than that of other related methods, again indicating the effectiveness of negative rationales.

\begin{table}[!ht]\small
    \renewcommand\arraystretch{1.15}
    \centering
    \caption{Results of our method on SVAMP with different sampling counts, i.e., the sampling count $m$ for positive rationales and $k$ for negative rationales.}
    \begin{tabular}{ccccc}
    \hline
        Positive sample m & 1 & 5 & 10 & 20 \\ \hline
        T5-base & 67.5 & 69.0 & 68.0 & 68.5 \\ \hline \hline
        negative sample k & 0 & 1 & 2 & 3 \\ \hline  
        T5-base & 67.0 & 69.0 & 68.5 & 66.0 \\ \hline
    \end{tabular}
    \label{table6}
\end{table}

\subsection{Rationality for negative knowledge}\label{temperature_sampling}
\begin{figure}[!ht]
 \centering
\includegraphics[width=1.\columnwidth]{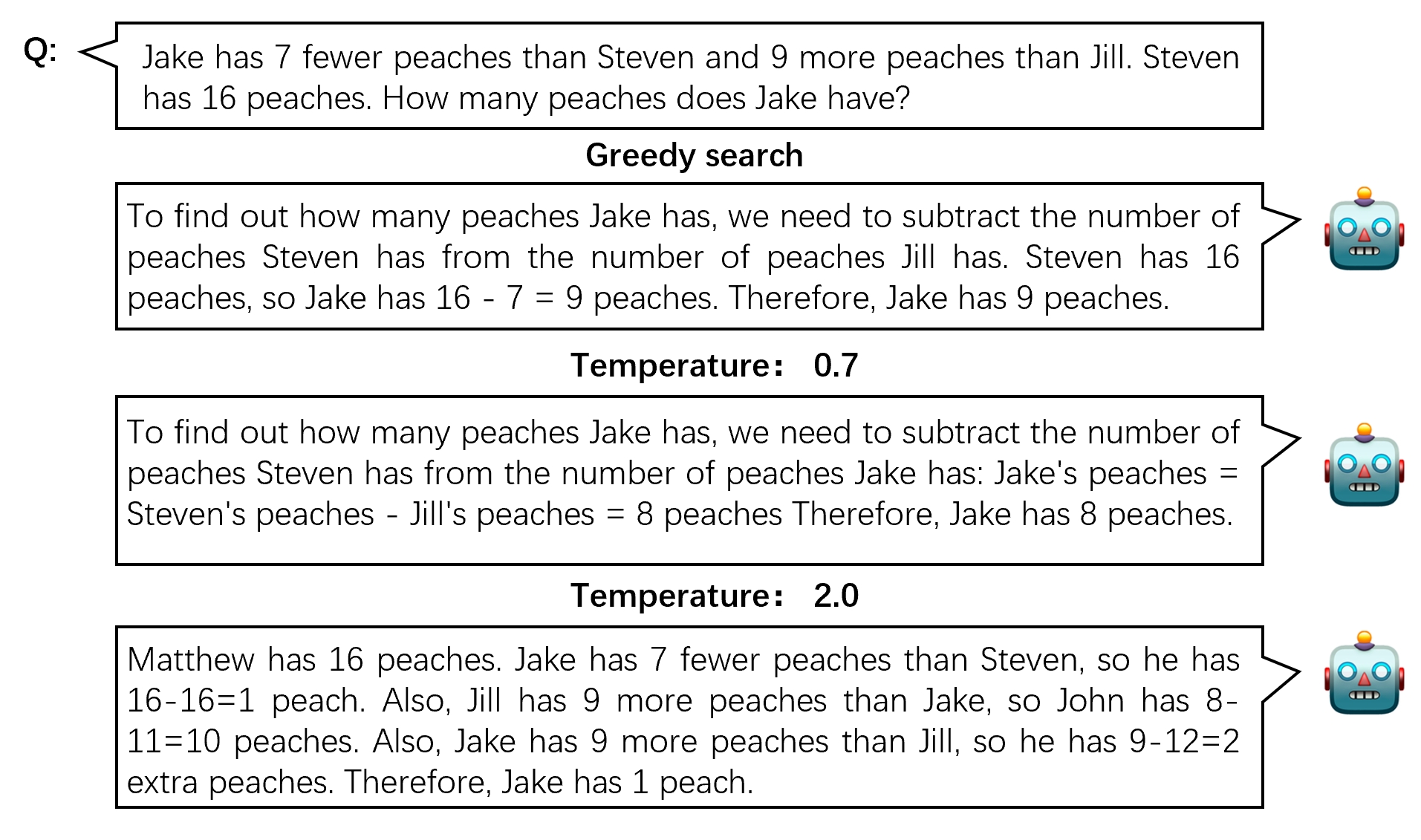}\\
\caption{A case study of output rationales from the T5-base model on SVAMP with different temperatures.}\label{sample}
\end{figure}

Temperature sampling is a commonly used decoding strategy for LLMs' generation process. By adjusting the temperature $\tau$, we can modify the probability distribution of each word before sampling. The higher the temperature is, the smaller the difference in the probability distribution of LLM's outputs becomes, increasing the chance of sampling words with lower probabilities. In Fig~\ref{sample}, we provided a case of output rationales from the trained T5-base model with different temperature settings for illustrative visualization. It validated the rationality that we generated negative rationales by sampling the iteration-before smaller models with a high temperature value. We further explored the influence of the negative sampling temperature on model performance in Appendix~\ref{temperature}. 

\subsection{Assessment for generated rationales}
We assessed the qualities of CoT examples using GPT-3.5-turbo. Inspired by the ranking model, we prompted GPT-3.5-turbo to rank the rationales generated by both DSS and our QCRD, rather than providing scores based on the qualities of the rationales. This is easier for the LLM and allows for a more straightforward comparison. The prompt fed to GPT-3.5-turbo was presented in Appendix Table \ref{prompt}. To evaluate the rationales, we randomly selected $50$ examples from each of the four datasets and asked GPT-3.5-turbo to determine rationales of which our method was better. We then aggregated the counts of "DSS is better," "Both are good," and "QCRD is better," as shown in Table~\ref{assessment}. From the results, we observed that on SVAMP, CQA, and ESNLI, the model trained using our method generated better rationales than using DSS. However, on ANLI1, the model trained using DSS exhibited slightly better performance.

\begin{table}[!ht]\small
    \renewcommand\arraystretch{1.25}
    \centering
    \caption{The quality assessment results on the T5-base model for different sampling temperature settings, where three numbers represent counts of "DSS is better", "Both are good", and "QCRD is better", respectively.}
    \setlength{\tabcolsep}{1.5mm}{
    \begin{tabular}{ccccc}
    \hline
         & SVAMP & CQA & ANLI1 & ESNLI \\ \hline
        $\tau=0$ & 21/0/29 & 19/6/25 & 26/1/23 & 17/11/22 \\
         $\tau=0.7$ &
          22/0/28 & 20/6/24 & 25/1/24 & 14/3/33 \\ \hline
    \end{tabular}}
    \label{assessment}
\end{table}

\subsection{Distribution of rationale quality scores}
\begin{figure}[!ht]
 \centering
\includegraphics[width=1\columnwidth]{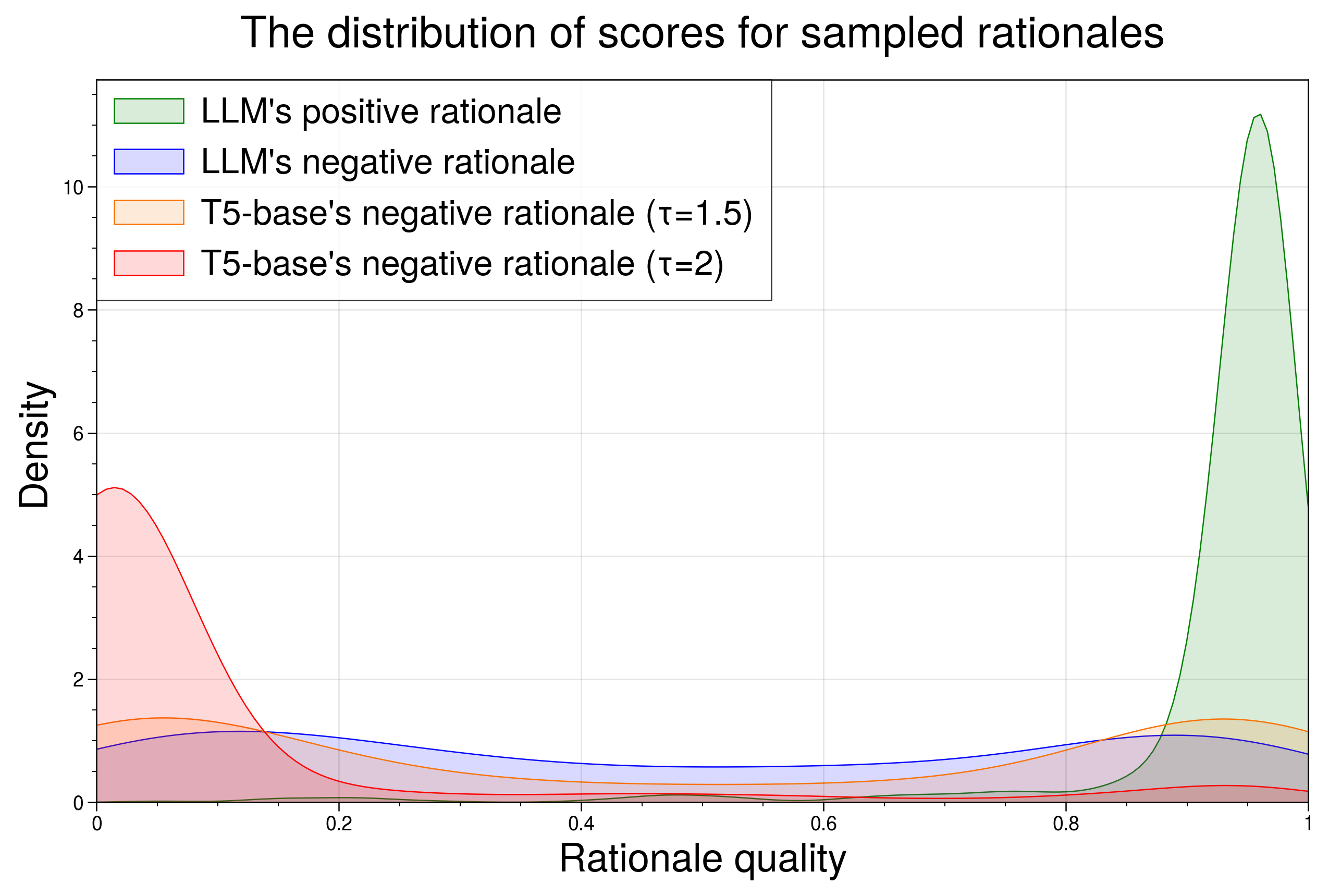}\\
\caption{The probability density estimation for sampled rationale scores on the SVAMP test dataset, where rationales were from LLM's and trained T5-base model's sampled outputs, and $\tau$ denotes sampling temperature.}\label{distribution}
\end{figure}

The probability density estimation for the sampled rationale scores from the trained discriminator on the SVAMP test dataset is shown in Fig~\ref{distribution}. Specifically, we considered the quality scores of: (1) positive and negative rationales from LLM's sampled outputs (sampled 5 times); (2) negative rationales from sampling a trained T5-base model with temperature $\tau$ set to 1.5 and 2.0, respectively. It showed that the trained discriminator can effectively score different types of rationales. Scores of LLM's positive rationales were around 0.95. For the trained student model, scores of the sampled negative rationales sometimes exceeded 0.7 (see the orange distribution), and it was necessary for the discriminator to assign low weights to these rationales. Furthermore, by comparing the orange distribution and the red distribution, we can see that the sampling temperature has a significant influence on the qualities of the rationales.

\section{Conclusion}
The knowledge distillation of CoT rationales from LLMs into smaller language models using a multi-task learning framework has been empirically shown to enhance performances of smaller language models. Building upon the framework, we introduces a general CoT distillation method, incorporating a contrastive learning perspective that considers both positive and negative knowledge. To generate positive and negative rationales, we propose an innovative approach that combines temperature sampling, the self-consistency of LLMs, and the self-adversarial of small language models themselves. Additionally, we develop a many-to-one contrastive distillation loss for better knowledge learning, where an online-update discriminator is used to judge qualities of rationales and assign them weights for optimizing the training process across the whole datasets. Extensive experiments conducted on multiple reasoning tasks demonstrate the superiority of our method over previous ones.

\section*{Limitations}
Our work serves as a distillation method for language models, paving the way for further improvements. On one hand, as illustrated in Appendix~\ref{training cost}, it requires additional training time due to the distillation of sampled positive and online-inferenced negative rationales, even though our proposed method can enhance model performance without incurring additional deployment costs. On the other hand, the quality of knowledge for distillation is crucial. In this paper, we prompt the LLM to generate CoT rationales and further classify them into positive and negative ones through self-consistency. However, different types of prompts and decoding strategies can lead the LLM to produce various forms of positive CoT knowledge and more intuitive negative CoT knowledge, which may further improve the distillation effect.
\bibliography{custom}

\begin{thebibliography}{30}
\providecommand{\natexlab}[1]{#1}

\bibitem[{Camburu et~al.(2018)Camburu, Rockt{\"a}schel, Lukasiewicz, and Blunsom}]{camburu2018snli}
Oana-Maria Camburu, Tim Rockt{\"a}schel, Thomas Lukasiewicz, and Phil Blunsom. 2018.
\newblock e-snli: Natural language inference with natural language explanations.
\newblock \emph{Advances in Neural Information Processing Systems}, 31.

\bibitem[{Chen et~al.(2024{\natexlab{a}})Chen, Saha, Stengel-Eskin, and Bansal}]{chen2024magdi}
Justin Chih-Yao Chen, Swarnadeep Saha, Elias Stengel-Eskin, and Mohit Bansal. 2024{\natexlab{a}}.
\newblock Magdi: Structured distillation of multi-agent interaction graphs improves reasoning in smaller language models.
\newblock \emph{arXiv preprint arXiv:2402.01620}.

\bibitem[{Chen et~al.(2024{\natexlab{b}})Chen, Huang, Gao, Wang, Zhao, and Ding}]{chen2024learning}
Xin Chen, Hanxian Huang, Yanjun Gao, Yi~Wang, Jishen Zhao, and Ke~Ding. 2024{\natexlab{b}}.
\newblock Learning to maximize mutual information for chain-of-thought distillation.
\newblock \emph{arXiv preprint arXiv:2403.03348}.

\bibitem[{Chowdhery et~al.(2023)Chowdhery, Narang, Devlin, Bosma, Mishra, Roberts, Barham, Chung, Sutton, Gehrmann et~al.}]{chowdhery2023palm}
Aakanksha Chowdhery, Sharan Narang, Jacob Devlin, Maarten Bosma, Gaurav Mishra, Adam Roberts, Paul Barham, Hyung~Won Chung, Charles Sutton, Sebastian Gehrmann, et~al. 2023.
\newblock Palm: Scaling language modeling with pathways.
\newblock \emph{Journal of Machine Learning Research}, 24(240):1--113.

\bibitem[{Fu et~al.(2023)Fu, Peng, Ou, Sabharwal, and Khot}]{fu2023specializing}
Yao Fu, Hao Peng, Litu Ou, Ashish Sabharwal, and Tushar Khot. 2023.
\newblock Specializing smaller language models towards multi-step reasoning.
\newblock In \emph{International Conference on Machine Learning}, pages 10421--10430. PMLR.

\bibitem[{Hinton et~al.(2015)Hinton, Vinyals, and Dean}]{hinton2015distilling}
Geoffrey Hinton, Oriol Vinyals, and Jeff Dean. 2015.
\newblock Distilling the knowledge in a neural network.
\newblock \emph{arXiv preprint arXiv:1503.02531}.

\bibitem[{Howard and Ruder(2018)}]{howard2018universal}
Jeremy Howard and Sebastian Ruder. 2018.
\newblock Universal language model fine-tuning for text classification.
\newblock \emph{arXiv preprint arXiv:1801.06146}.

\bibitem[{Hsieh et~al.(2023)Hsieh, Li, Yeh, Nakhost, Fujii, Ratner, Krishna, Lee, and Pfister}]{hsieh2023distilling}
Cheng-Yu Hsieh, Chun-Liang Li, Chih-Kuan Yeh, Hootan Nakhost, Yasuhisa Fujii, Alexander Ratner, Ranjay Krishna, Chen-Yu Lee, and Tomas Pfister. 2023.
\newblock Distilling step-by-step! outperforming larger language models with less training data and smaller model sizes.
\newblock \emph{arXiv preprint arXiv:2305.02301}.

\bibitem[{Jaiswal et~al.(2020)Jaiswal, Babu, Zadeh, Banerjee, and Makedon}]{jaiswal2020survey}
Ashish Jaiswal, Ashwin~Ramesh Babu, Mohammad~Zaki Zadeh, Debapriya Banerjee, and Fillia Makedon. 2020.
\newblock A survey on contrastive self-supervised learning.
\newblock \emph{Technologies}, 9(1):2.

\bibitem[{Kim and Rush(2016)}]{kim2016sequence}
Yoon Kim and Alexander~M Rush. 2016.
\newblock Sequence-level knowledge distillation.
\newblock \emph{arXiv preprint arXiv:1606.07947}.

\bibitem[{Kojima et~al.(2022)Kojima, Gu, Reid, Matsuo, and Iwasawa}]{kojima2022large}
Takeshi Kojima, Shixiang~Shane Gu, Machel Reid, Yutaka Matsuo, and Yusuke Iwasawa. 2022.
\newblock Large language models are zero-shot reasoners.
\newblock \emph{Advances in neural information processing systems}, 35:22199--22213.

\bibitem[{Le-Khac et~al.(2020)Le-Khac, Healy, and Smeaton}]{le2020contrastive}
Phuc~H Le-Khac, Graham Healy, and Alan~F Smeaton. 2020.
\newblock Contrastive representation learning: A framework and review.
\newblock \emph{Ieee Access}, 8:193907--193934.

\bibitem[{Li et~al.(2022)Li, Chen, Shen, Chen, Zhang, Li, Wang, Qian, Peng, Mao et~al.}]{li2022explanations}
Shiyang Li, Jianshu Chen, Yelong Shen, Zhiyu Chen, Xinlu Zhang, Zekun Li, Hong Wang, Jing Qian, Baolin Peng, Yi~Mao, et~al. 2022.
\newblock Explanations from large language models make small reasoners better.
\newblock \emph{arXiv preprint arXiv:2210.06726}.

\bibitem[{Li et~al.(2024)Li, Yuan, Feng, Pan, Sun, Wang, Wang, and Li}]{li2024turning}
Yiwei Li, Peiwen Yuan, Shaoxiong Feng, Boyuan Pan, Bin Sun, Xinglin Wang, Heda Wang, and Kan Li. 2024.
\newblock Turning dust into gold: Distilling complex reasoning capabilities from llms by leveraging negative data.
\newblock In \emph{Proceedings of the AAAI Conference on Artificial Intelligence}, volume~38, pages 18591--18599.

\bibitem[{Magister et~al.(2022)Magister, Mallinson, Adamek, Malmi, and Severyn}]{magister2022teaching}
Lucie~Charlotte Magister, Jonathan Mallinson, Jakub Adamek, Eric Malmi, and Aliaksei Severyn. 2022.
\newblock Teaching small language models to reason.
\newblock \emph{arXiv preprint arXiv:2212.08410}.

\bibitem[{Nie et~al.(2019)Nie, Williams, Dinan, Bansal, Weston, and Kiela}]{nie2019adversarial}
Yixin Nie, Adina Williams, Emily Dinan, Mohit Bansal, Jason Weston, and Douwe Kiela. 2019.
\newblock Adversarial nli: A new benchmark for natural language understanding.
\newblock \emph{arXiv preprint arXiv:1910.14599}.

\bibitem[{Patel et~al.(2021)Patel, Bhattamishra, and Goyal}]{patel2021nlp}
Arkil Patel, Satwik Bhattamishra, and Navin Goyal. 2021.
\newblock Are nlp models really able to solve simple math word problems?
\newblock \emph{arXiv preprint arXiv:2103.07191}.

\bibitem[{Phuong and Lampert(2019)}]{phuong2019towards}
Mary Phuong and Christoph Lampert. 2019.
\newblock Towards understanding knowledge distillation.
\newblock In \emph{International conference on machine learning}, pages 5142--5151. PMLR.

\bibitem[{Raffel et~al.(2020)Raffel, Shazeer, Roberts, Lee, Narang, Matena, Zhou, Li, and Liu}]{raffel2020exploring}
Colin Raffel, Noam Shazeer, Adam Roberts, Katherine Lee, Sharan Narang, Michael Matena, Yanqi Zhou, Wei Li, and Peter~J Liu. 2020.
\newblock Exploring the limits of transfer learning with a unified text-to-text transformer.
\newblock \emph{Journal of machine learning research}, 21(140):1--67.

\bibitem[{Renze and Guven(2024)}]{renze2024effect}
Matthew Renze and Erhan Guven. 2024.
\newblock The effect of sampling temperature on problem solving in large language models.
\newblock \emph{arXiv preprint arXiv:2402.05201}.

\bibitem[{Silver et~al.(2018)Silver, Hubert, Schrittwieser, Antonoglou, Lai, Guez, Lanctot, Sifre, Kumaran, Graepel et~al.}]{silver2018general}
David Silver, Thomas Hubert, Julian Schrittwieser, Ioannis Antonoglou, Matthew Lai, Arthur Guez, Marc Lanctot, Laurent Sifre, Dharshan Kumaran, Thore Graepel, et~al. 2018.
\newblock A general reinforcement learning algorithm that masters chess, shogi, and go through self-play.
\newblock \emph{Science}, 362(6419):1140--1144.

\bibitem[{Stanovich and West(1991)}]{stanovich199124}
Keith~E Stanovich and Richard~F West. 1991.
\newblock 24. individual differences in reasoning: Implications for the rationality debate?

\bibitem[{Talmor et~al.(2018)Talmor, Herzig, Lourie, and Berant}]{talmor2018commonsenseqa}
Alon Talmor, Jonathan Herzig, Nicholas Lourie, and Jonathan Berant. 2018.
\newblock Commonsenseqa: A question answering challenge targeting commonsense knowledge.
\newblock \emph{arXiv preprint arXiv:1811.00937}.

\bibitem[{Wang et~al.(2023)Wang, Wang, Li, Gao, Yin, and Ren}]{wang2023scott}
Peifeng Wang, Zhengyang Wang, Zheng Li, Yifan Gao, Bing Yin, and Xiang Ren. 2023.
\newblock Scott: Self-consistent chain-of-thought distillation.
\newblock \emph{arXiv preprint arXiv:2305.01879}.

\bibitem[{Wang et~al.(2022)Wang, Wei, Schuurmans, Le, Chi, Narang, Chowdhery, and Zhou}]{wang2022self}
Xuezhi Wang, Jason Wei, Dale Schuurmans, Quoc Le, Ed~Chi, Sharan Narang, Aakanksha Chowdhery, and Denny Zhou. 2022.
\newblock Self-consistency improves chain of thought reasoning in language models.
\newblock \emph{arXiv preprint arXiv:2203.11171}.

\bibitem[{Wei et~al.(2022{\natexlab{a}})Wei, Tay, Bommasani, Raffel, Zoph, Borgeaud, Yogatama, Bosma, Zhou, Metzler et~al.}]{wei2022emergent}
Jason Wei, Yi~Tay, Rishi Bommasani, Colin Raffel, Barret Zoph, Sebastian Borgeaud, Dani Yogatama, Maarten Bosma, Denny Zhou, Donald Metzler, et~al. 2022{\natexlab{a}}.
\newblock Emergent abilities of large language models.
\newblock \emph{arXiv preprint arXiv:2206.07682}.

\bibitem[{Wei et~al.(2022{\natexlab{b}})Wei, Wang, Schuurmans, Bosma, Xia, Chi, Le, Zhou et~al.}]{wei2022chain}
Jason Wei, Xuezhi Wang, Dale Schuurmans, Maarten Bosma, Fei Xia, Ed~Chi, Quoc~V Le, Denny Zhou, et~al. 2022{\natexlab{b}}.
\newblock Chain-of-thought prompting elicits reasoning in large language models.
\newblock \emph{Advances in neural information processing systems}, 35:24824--24837.

\bibitem[{Zelikman et~al.(2022)Zelikman, Wu, Mu, and Goodman}]{zelikman2022star}
Eric Zelikman, Yuhuai Wu, Jesse Mu, and Noah Goodman. 2022.
\newblock Star: Bootstrapping reasoning with reasoning.
\newblock \emph{Advances in Neural Information Processing Systems}, 35:15476--15488.

\bibitem[{Zeman et~al.(2018)Zeman, Hajic, Popel, Potthast, Straka, Ginter, Nivre, and Petrov}]{zeman2018conll}
Daniel Zeman, Jan Hajic, Martin Popel, Martin Potthast, Milan Straka, Filip Ginter, Joakim Nivre, and Slav Petrov. 2018.
\newblock Conll 2018 shared task: Multilingual parsing from raw text to universal dependencies.
\newblock In \emph{Proceedings of the CoNLL 2018 Shared Task: Multilingual parsing from raw text to universal dependencies}, pages 1--21.

\bibitem[{Zhu et~al.(2023)Zhu, Li, Liu, Ma, and Wang}]{zhu2023survey}
Xunyu Zhu, Jian Li, Yong Liu, Can Ma, and Weiping Wang. 2023.
\newblock A survey on model compression for large language models.
\newblock \emph{arXiv preprint arXiv:2308.07633}.

\end{thebibliography}

\appendix

\section{Appendix}\label{sec:appendix}
\setcounter{figure}{0}
\setcounter{table}{0}

\subsection{Details about Datasets}
\label{annotations}
Following the setting in~\cite{hsieh2023distilling}, we provide detailed descriptions of the four benchmark datasets in Table~\ref{datasets}. To illustrate the unbalanced proportion of positive and negative rationales from LLMs given the ground truth, we displayed the statistical description of the generated rationale annotations on training datasets for four benchmarks in Table~\ref{annotation}. On the one hand, the number of positive rationales was larger than that of negative rationales (3.87:1.13). On the other hand, for many samples in the training dataset (more than $50\%$), there were only positive rationales. Therefore, there is a need to generate effective negative rationales in other ways.

\begin{table}[!ht]\small
    \centering
    \renewcommand\arraystretch{1.2}
    \caption{Descriptions of the four benchmark datasets.}
    \begin{tabular}{cccc}
    \hline
        Dataset & Training & Validation & Test \\ \hline
        SVAMP & 720 & 80  & 200 \\
        CQA & 8766 & 975 & 1221 \\
        ANLI1 & 16946 & 1000 & 1000 \\
        ESNLI & 549367 & 9842 & 9824 \\ \hline
    \end{tabular}
    \label{datasets}
\end{table}

\begin{table}[!ht]\small
    \centering
    \renewcommand\arraystretch{1.2}
    \setlength{\tabcolsep}{1.mm}{
    \caption{Statistical descriptions of the generated rationale annotations, where r denotes rationale, and positive r achieves correct answers.}
    \begin{tabular}{ccccc}
    \hline
        Dataset & SVAMP & CQA & ANLI1 & ESNLI \\ \hline
        Average pos r (total 5) & 3.87 & 3.89  & 3.93 & 3.31 \\
        Proportion with only pos r & 0.55 & 0.68 & 0.66 & 0.50 \\
        Proportion with only neg r & 0.08 & 0.13 & 0.11 & 0.20 \\ \hline
    \end{tabular}
    \label{annotation}}
\end{table}

\subsection{Iteration-before-models for negative rationale generators}
\label{Iteration-before-models}
\begin{figure}[!ht]\small
\centering
\includegraphics[width=0.80\columnwidth]{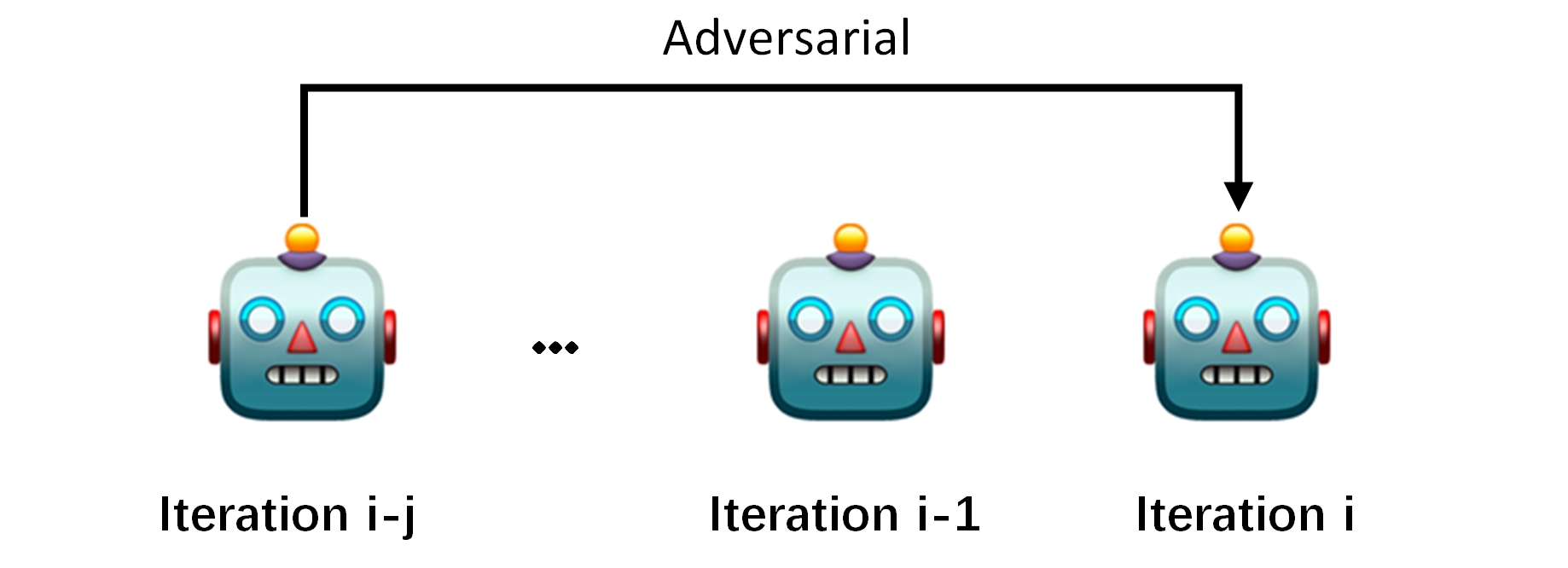}\\
\caption{A case of the j-iteration-before-model for the negative rationale generator.}\label{appendix}
\end{figure}

In this paper, we dynamically generated negative rationales using iteration-before-models through online temperature sampling. We took these iteration-before-models as negative generators, and we sampled them with a relatively high temperature value to generate negative rationales for every batch of datasets. As depicted in Fig.~\ref{appendix}, to select a j-iteration-before-model for the negative rationale generator, we need to save a minimum of $j$ checkpoints for the model. This allows us to load the negative generator online and train the student model end-to-end instead of using a multi-turn approach. Additionally, as shown in Table~\ref{negativeways}, we found that the performance of the student models was not sensitive to the choice of $j$ from $\{3, 5, 10\}$, all of which outperformed the results obtained with a fixed negative generator (pretrained by DSS~\cite{hsieh2023distilling}) or by using negative rationales derived from the self-consistency of the LLM.

\begin{table}[!ht]\small
    \renewcommand\arraystretch{1.25}
    \centering
    \caption{Results on SVAMP with different negative knowledge strategies, where the "Fixed" denotes fixing negative generator with the pretrained model, "SC" denotes using negative rationales from self-consistency.}
    \begin{tabular}{cccccc}
    \hline
        Negative source & $j$=3 & $j$=5 & $j$=10 & Fixed & SC\\ \hline
        T5-base & 68.0 & 69.0 & 68.5 & 66.5 & 64.5 \\ \hline
    \end{tabular}
    \label{negativeways}
\end{table}

\begin{table}[!ht]\small
    \renewcommand\arraystretch{1.5}
    \centering
    \caption{The prompt for GPT-3.5-turbo to judge rationales.}
    \begin{tabular}{p{7cm}}
    \hline
        \multicolumn{1}{c}{The prompt for GPT-3.5-turbo}  \\ \hline
        There is an input pair of a question and an answer of a taskname task, and we provide you two explanations. You need judge which explanation is better. The better explanation should be more accurate and explain the answer better.   \\ \hline
    \end{tabular}
    \label{prompt}
\end{table}

\subsection{Details about the Quality Judge}\label{Judge}
We incorporated a discriminator into our training process to assess the quality of rationales and assign corresponding weights to the losses. To construct the discriminator, we leveraged the encoder of the T5-base model along with one maxpooling layer and two linear layers to compute the quality score. Prior to training, the discriminator needs to be pretrained using the output rationales generated by LLMs with applying data augmentations to the negative rationales. Specifically, we employed word mask and replacement with the assistance of StanfordNLP~\cite{zeman2018conll} to balance the proportions of positive and negative rationales. The training objective is $L_{D}$ in (\ref{discriminatorloss}). We pretrained the discriminator 500 max steps and we ensured scores for positive rationales close to $1$ and scores for negative rationales close to $0$. The discriminator was further online-updated during training. 

\subsection{Influence of negative sampling temperature}\label{temperature}
The results of the T5-base model distilled by our method on SVAMP were displayed in Table~\ref{sampling} in terms of different negative sampling temperature settings. It was observed that when no sampling was performed (i.e., $\tau=0$) or a lower temperature value was used (i.e., $\tau=0.7$), the smaller model exhibited relatively poorer performance and showed larger fluctuations in accuracy. The best results were achieved when the temperature $\tau$ was set to $1.5$. The reason for this can be attributed to the fact that when the model approaches convergence, the output rationales with lower temperature values tend to be similar to the golden ones. Considering these similar outputs as negative samples can lead to detrimental effects.

\begin{table}[!ht]\small
    \renewcommand\arraystretch{1.25}
    \centering
    \caption{Results of our method on SVAMP with different sampling temperature $\tau$.}
    \begin{tabular}{ccccc}
    \hline
        Temperature $\tau$ & 0 & 0.7 & 1.5 & 2 \\ \hline
        T5-base & 65.0 & 64.5 & 69.0 & 67.5 \\ \hline
    \end{tabular}
    \label{sampling}
\end{table}

\subsection{Computational cost}\label{training cost}
The training times for the T5-base and T5-small models using each method are shown in Table~\ref{cost}. Specifically, we trained the models on a single A100-80G GPU using the SVAMP benchmark. Compared to DSS~\cite{hsieh2023distilling}, our method requires an additional 9 hours for T5-base and 6 hours for T5-small. This increase is due to each input necessitating 5 positive rationales and 1 online-inferenced negative rationale for contrastive rationale distillation. However, we emphasize the motivation behind our method: to enhance the performance of deployed small language models. Our proposed QCRD effectively improves model performance without incurring additional parameter storage during deployment.

\begin{table}[!ht]\small
    \renewcommand\arraystretch{1.25}
    \centering
    \caption{The comparison for training time on T5-base and T5-small models, where D dnotes the Quality Judge.}
    \begin{tabular}{cc}
    \hline
        Method &  Base/small training time (h) \\ \hline
        Finetune & 2.0 / 1.25 \\ 
        DSS & 4.0 / 2.5 \\ 
        MI & 4.2 / 2.6 \\ 
        QCRD & 13.0 / 8.5 \\
        QCRD (w/o D) & 12.0 / 7.5 \\ \hline
        
    \end{tabular}
    \label{cost}
\end{table}

\end{document}